\pdfoutput=1
\documentclass[12pt, final]{l4dc2023}

\usepackage{times}
\usepackage{caption}
\usepackage{fixltx2e}
\usepackage{wrapfig}
\usepackage{multirow}
\usepackage{enumitem}
\usepackage{xfrac}
\usepackage{float}

\title[Physics-Informed Model-Based Reinforcement Learning]{Physics-Informed Model-Based Reinforcement Learning}
\usepackage{times}
\author{%
\Name{Adithya Ramesh}\\
\addr Robert Bosch Centre for Data Science and Artificial Intelligence, Indian Institute of Technology Madras
\AND
\Name{Balaraman Ravindran}\\
\addr Robert Bosch Centre for Data Science and Artificial Intelligence, Indian Institute of Technology Madras\\
\addr Department of Computer Science and Engineering, Indian Institute of Technology Madras
}

\begin{document}

\maketitle

\begin{abstract}%
We apply reinforcement learning (RL) to robotics tasks. One of the drawbacks of traditional RL algorithms has been their poor sample efficiency. One approach to improve the sample efficiency is model-based RL. In our model-based RL algorithm, we learn a model of the environment, essentially its transition dynamics and reward function, use it to generate imaginary trajectories and backpropagate through them to update the policy, exploiting the differentiability of the model. Intuitively, learning more accurate models should lead to better model-based RL performance. Recently, there has been growing interest in developing better deep neural network based dynamics models for physical systems, by utilizing the structure of the underlying physics. We focus on robotic systems undergoing rigid body motion without contacts. We compare two versions of our model-based RL algorithm, one which uses a standard deep neural network based dynamics model and the other which uses a much more accurate, physics-informed neural network based dynamics model. We show that, in model-based RL, model accuracy mainly matters in environments that are sensitive to initial conditions, where numerical errors accumulate fast. In these environments, the physics-informed version of our algorithm achieves significantly better average-return and sample efficiency. In environments that are not sensitive to initial conditions, both versions of our algorithm achieve similar average-return, while the physics-informed version achieves better sample efficiency. We also show that, in challenging environments, physics-informed model-based RL achieves better average-return than state-of-the-art model-free RL algorithms such as Soft Actor-Critic, as it computes the policy-gradient analytically, while the latter estimates it through sampling.
\end{abstract}

\begin{keywords}%
    Model-Based Reinforcement Learning, Robotics, Physics-Informed Neural Networks
\end{keywords}

\section{Introduction}
We apply reinforcement learning (RL) to robotics tasks. RL can solve sequential decision making problems through trial and error. In recent years, RL has been combined with powerful function approximators such as deep neural networks to solve complex robotics tasks with high-dimensional state and action spaces~\citep{ddpg,ppo,td3,sacv2}. However, there remain some critical challenges in RL today~\citep{real-world-rl}. One of them is sample efficiency. Traditional RL algorithms require a lot of interactions with the environment to learn successful policies. In robotics, collecting such large amounts of training data using actual robots is not practical. One solution is to train in simulation. However, for many robotics tasks, developing realistic simulations is hard. Another solution is to improve the sample efficiency. One approach to improve the sample efficiency is model-based RL. Here, we learn a model of the environment, essentially its transition dynamics and reward function, use it to generate imaginary trajectories and use that data to update the policy. This way we need fewer interactions with the actual environment. Some model-based RL algorithms use the model just to generate additional data and update the policy using a model-free algorithm, for e. g., ~\citealp{sutton1991dyna,janner2019trust}. Other model-based RL algorithms exploit the differentiability of the model and backpropagate through the imaginary trajectories to update the policy, for e. g., ~\citealp{deisenroth2011pilco,heess2015learning,clavera2020model,hafner2019dream,hafner2020mastering}. We adopt the second approach.

Intuitively, learning more accurate models should lead to better model-based RL performance. Recently, there has been growing interest in developing better deep neural network based dynamics models for physical systems, by utilizing the structure of the underlying physics~\citep{lutter2019deep,hnn,lutter2019deepenergy,zhong2019symplectic,zhong2020dissipative,lnn,finzi2020simplifying,zhong2021benchmarking}. These physics-informed neural networks learn the dynamics much more accurately compared to standard deep neural networks. They also obey the underlying physical laws, such as conservation of energy much better. Previous studies in this area have mostly focused on improving dynamics learning. Some studies have learnt a physics-informed neural network based dynamics model and used it for downstream tasks such as inverse dynamics control~\citep{lutter2019deep} and energy based control~\citep{lutter2019deepenergy,zhong2019symplectic}. We use it to train a model-based RL algorithm.

We focus on robotic systems undergoing rigid body motion without contacts. We compare two versions of our model-based RL algorithm, one which uses a standard deep neural network based dynamics model and the other which uses a much more accurate, physics-informed neural network based dynamics model. Our contributions are as follows,
\begin{itemize}[leftmargin=*]
\item Our first contribution is using a physics-informed neural network based dynamics model to train a model-based RL algorithm. We are one of the first to do so.

\item Our second contribution is showing that, in model-based RL, model accuracy mainly matters in environments that are sensitive to initial conditions, where numerical errors accumulate fast. In these environments, the physics-informed version of our algorithm achieves significantly better average-return and sample efficiency. In environments that are not sensitive to initial conditions, both versions of our algorithm achieve similar average-return, while the physics-informed version achieves better sample efficiency.

\item The sensitivity to initial conditions depends on factors such as the system dynamics, degree of actuation, control policy and damping. We measure it using the finite-time maximal Lyapunov exponent. We compute the same using the variational equation~\citep{skokos2010lyapunov}, which linearizes the dynamics to estimate how separation vectors evolve with time. The standard variational equation is defined only for unforced autonomous systems. We extend it to forced autonomous systems. This is our third contribution.

\item Our fourth contribution is showing that, in challenging environments, physics-informed model-based RL achieves better average-return than state-of-the-art model-free RL algorithms such as Soft Actor-Critic~\citep{sacv2}, as it computes the policy-gradient analytically, while the latter estimates it through sampling.
\end{itemize}

\section{Environments}
\begin{figure}[h]
    \centering
    % \captionsetup{justification=centering,width=.95\linewidth}
    % \captionsetup{width=.8\linewidth}
    \subfigure[\scriptsize Reacher]{\includegraphics[width=0.12\textwidth]{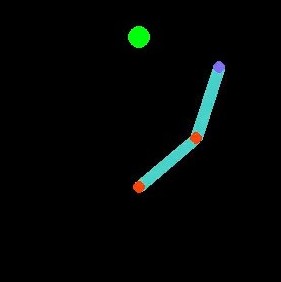}}
    \subfigure[\scriptsize Pendulum]{\includegraphics[width=0.12\textwidth]{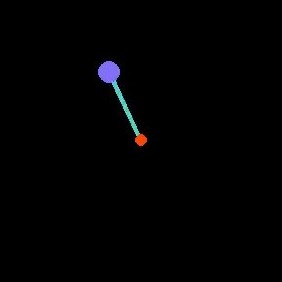}}
    \subfigure[\scriptsize Cartpole]{\includegraphics[width=0.12\textwidth]{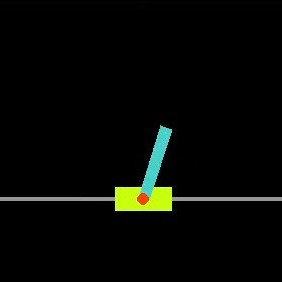}}
    \subfigure[\scriptsize Cart-2-pole]{\includegraphics[width=0.122\textwidth]{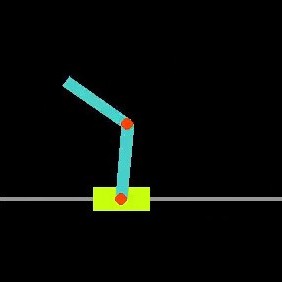}}
    \subfigure[\scriptsize Acrobot]{\includegraphics[width=0.12\textwidth]{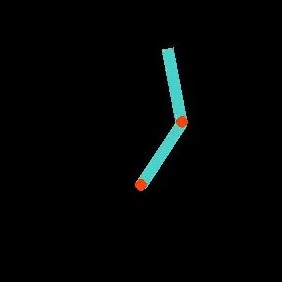}}
    \subfigure[\scriptsize Cart-3-pole]{\includegraphics[width=0.12\textwidth]{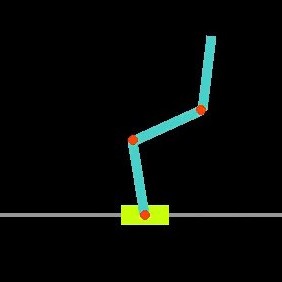}}
    \subfigure[\scriptsize Acro-3-bot]{\includegraphics[width=0.12\textwidth]{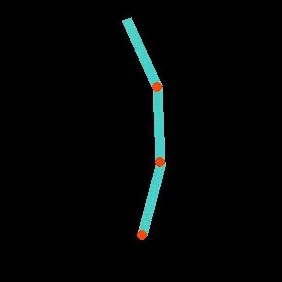}}
    \caption{We consider robotic systems undergoing rigid body motion without contacts.}
   \label{fig:envs}
\end{figure}

We focus on robotic systems undergoing rigid body motion without contacts. We also assume that there is no friction. The environments considered are shown in Figure \ref{fig:envs}. We develop our own simulations from first principles. We provide more details about the environments, such as the task to be accomplished, which joints are acutated, etc, in Appendix \ref{sec:env}. In future work, we plan to include both friction as well as contacts.

\subsection{Lagrangian Mechanics}
These systems obey Lagrangian mechanics. Their state consists of generalized coordinates $\textbf{q}$, which describe the configuration of the system, and generalized velocities $\dot{\textbf{q}}$, which are the time derivatives of $\textbf{q}$. The Lagrangian is a scalar quantity defined as $\mathcal{L}(\textbf{q},\dot{\textbf{q}},t) = \mathcal{T}(\textbf{q}, \dot{\textbf{q}})-\mathcal{V}(\textbf{q})$, where $\mathcal{T}(\textbf{q}, \dot{\textbf{q}})$ is the kinetic energy and $\mathcal{V}(\textbf{q})$ is the potential energy. Let the motor torques be $\boldsymbol\tau$. The Lagrangian equations of motion are given by,
\begin{equation} 
\dfrac{d}{dt}\dfrac{\partial \mathcal{L}}{\partial \dot{\textbf{q}}}-\dfrac{\partial \mathcal{L}}{\partial \textbf{q}} = \boldsymbol\tau 
\end{equation}

For rigid body motion, $\mathcal{T}(\textbf{q}, \dot{\textbf{q}}) = \frac{1}{2} \, \dot{\textbf{q}}^{T} \, \textbf{M}(\textbf{q}) \, \dot{\textbf{q}}$, where $\textbf{M}(\textbf{q})$ is the mass matrix, which is symmetric and positive definite. Hence, the Lagrangian equations of motion become,
\begin{equation}
\textbf{M}(\textbf{q}) \, \ddot{\textbf{q}} + \textbf{C}(\textbf{q},\dot{\textbf{q}}) \, \dot{\textbf{q}} + \textbf{G}(\textbf{q}) = \boldsymbol\tau
\label{eq-1}
\end{equation}
where, $\textbf{C}(\textbf{q},\dot{\textbf{q}}) \, \dot{\textbf{q}} = \frac{\partial }{\partial \textbf{q}} \big(\textbf{M}(\textbf{q})\, \dot{\textbf{q}} \big) \, \dot{\textbf{q}} - \frac{\partial }{\partial \textbf{q}} \big( \frac{1}{2} \, \dot{\textbf{q}}^{T} \, \textbf{M}(\textbf{q})\, \dot{\textbf{q}} \big)$, is the centripetal / Coriolis term and $\textbf{G}(\textbf{q}) = \frac{\partial \mathcal{V}(\textbf{q})}{\partial \textbf{q}}$, is the gravitational term. 

\section{Model-Based RL}
 Our model-based RL algorithm essentially iterates over three steps. First is the environment interaction step, where we use the current policy to interact with the environment and gather data. Second is the model learning step, where we use the gathered data to learn the dynamics and reward models. Third is the behaviour learning step, where we use the learned model to generate imaginary trajectories and backpropagate through them to update the policy, exploiting the differentiability of the model. We discuss the model learning and behaviour learning steps in detail below. 

\subsection{Model Learning}
In the model learning step, we learn the dynamics and reward models. In dynamics learning, we want to predict the next state, given the current state and action, i. e., we want to learn the transformation $(\textbf{q}_{t}, \dot{\textbf{q}}_{t},\boldsymbol\tau_{t}) \rightarrow (\textbf{q}_{t+1}, \dot{\textbf{q}}_{t+1})$. The most straightforward approach is to train a standard deep neural network. We refer to this approach as DNN. This is shown in Figure \ref{fig:dnn}. 

Another approach is to utilize the structure of the underlying Lagrangian mechanics. This approach builds upon recent work such as Deep Lagrangian Networks (DeLaN)~\citep{lutter2019deep}, DeLaN for energy control~\citep{lutter2019deepenergy} and Lagrangian Neural Networks~\citep{lnn}. We detail this approach here. We use one network to learn the potential energy function $\mathcal{V}(\textbf{q})$ and another network to learn a lower triangular matrix $\textbf{L}(\textbf{q})$, using which we compute the mass matrix as $\textbf{M}(\textbf{q}) = \textbf{L}(\textbf{q})\;\textbf{L}^{T}(\textbf{q})$. We then compute the centripetal / Coriolis term $\textbf{C}(\textbf{q},\dot{\textbf{q}}) \, \dot{\textbf{q}}$ and the gravitational term $\textbf{G}(\textbf{q})$. Then, by rearranging Equation \ref{eq-1}, we get the acceleration as $\ddot{\textbf{q}} = \textbf{M}^{-1}(\textbf{q})\,(\boldsymbol\tau - \textbf{C}(\textbf{q},\dot{\textbf{q}}) \, \dot{\textbf{q}} - \textbf{G}(\textbf{q}))$. We then numerically integrate the state derivative $(\dot{\textbf{q}}, \ddot{\textbf{q}})$ over one time step using second-order Runge-Kutta to compute the next state. We refer to this approach as LNN, short for Lagrangian Neural Network. The entire process is shown in Figure \ref{fig:lnn}. In both the DNN and LNN approaches to dynamics learning, we use the L1 error between the predicted state and the ground truth as the loss for training.

In reward learning, we want to learn the reward function. In general, the reward is a function of the current state, action and the next state. In our case, the reward only depends on the next state. Hence, we train a network to map the next state to the reward. We use the L1 error between the predicted reward and the ground truth as the loss for training.
\begin{center}
\begin{minipage}{\linewidth}\centering
  \begin{minipage}{0.15\linewidth}\centering
      \begin{figure}[H]
          \includegraphics[width=\linewidth]{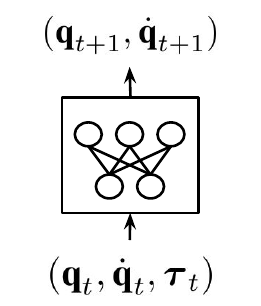}
          \captionsetup{justification=centering}
          \caption{DNN}
          \label{fig:dnn}
      \end{figure}
  \end{minipage}
  \hspace{0.1\linewidth}
  \begin{minipage}{0.48\linewidth}\centering
      \begin{figure}[H]
          \includegraphics[width=\textwidth]{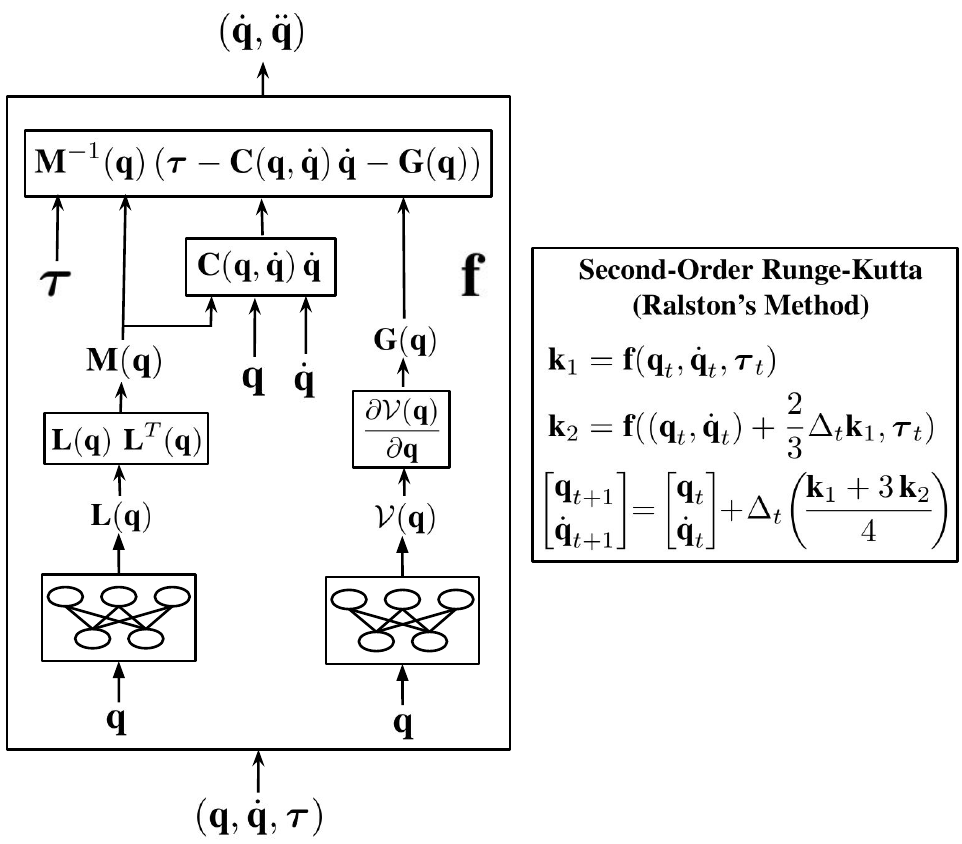}
          \captionsetup{justification=centering}
          \caption{LNN}
           \label{fig:lnn}
      \end{figure}
  \end{minipage}
\end{minipage}
\end{center}

\subsection{Behaviour Learning}
In the behaviour learning step, we use the learned model to generate imaginary trajectories and backpropagate through them to update the policy, exploiting the differentiability of the model. We build upon the Dreamer algorithm~\citep{hafner2019dream,hafner2020mastering}. We adopt an actor-critic approach. The critic aims to predict the expected discounted return from a given state. We train the critic to regress the $\lambda$-return~\citep{gae} computed using a target network that is updated every 100 critic updates,
\begin{equation}
V'_{\lambda}(s_{t}) = \begin{cases} \; r_{t} + \gamma \, (\,
(1-\lambda)V'(s_{t+1};w') + \lambda V'_{\lambda}(s_{t+1}) \,) & \text{if} \; t < T\\
\; V'(s_{t};w') & \text{if} \; t = T
\end{cases}
\end{equation}
The critic loss function is given by $L(w)  = \mathbb{E}\:[\sum_{t=0}^{T-1} \frac{1}{2} ( \: V(s_{t};w) - \text{sg}\,(V'_{\lambda}(s_{t})) \: ) ^{2}]$. We stop the gradients around the target (denoted by the $\text{sg}(.)$ function), as is typical in the literature.

We use a stochastic actor. The actor aims to output actions that lead to states that maximize the expected discounted return. We train the actor to maximize the same $\lambda$-return that was computed to train the critic. We add an entropy term to the actor objective to encourage exploration. The overall actor loss function is given by $L(\theta)  = - \; \mathbb{E}\:[\sum_{t=0}^{T-1} [\:V'_{\lambda}(s_{t}) - \eta \log \pi (a_{t}|s_{t};\theta) \:]]$. To backpropagate through sampled actions, we use the reparameterization trick~\citep{kingma2013auto}. The actor network outputs the mean $\mu$ and standard deviation $\sigma$ of a Gaussian distribution, from which we obtain the action as $a_{t} = \text{tanh}(\mu_{\theta}(s_{t})+\sigma_{\theta}(s_{t})\cdot \epsilon)$, where $\epsilon \sim \mathcal{N}(0,\mathbb{I})$. We summarize our overall model-based RL algorithm in Algorithm \ref{algo-1}.
\begin{algorithm2e}[h]
\begingroup \fontsize{9pt}{10pt}\selectfont
\DontPrintSemicolon % Some LaTeX compilers require you to use \dontprintsemicolon instead
Initialize networks with random weights.\;
Execute random actions for $K$ episodes to initialize replay buffer.\;
\For{each episode}{
    // \textbf{Model Learning}\;
    \For{$N_{1}$ times}{
    Draw mini batch of transitions from replay buffer. Fit dynamics and reward models.\;
    }
    // \textbf{Behaviour Learning}\;
    \For{$N_{2}$ times}{
    Draw mini batch of states from replay buffer. From each state, imagine a trajectory of length $T$.\;
    Construct actor, critic losses. Backpropagate through imaginary trajectories to update actor, critic.\;
    Every 100 updates, copy critic weights into critic target.\;
    }
    // \textbf{Environment Interaction}\;
    Interact with environment for one episode using current policy. Add transitions to replay buffer.\;
}
\endgroup
\caption{{Model-Based RL Algorithm.}}
\label{algo-1}
\end{algorithm2e}

\subsection{Experiments}
We train two versions of our model-based RL algorithm, one which uses the DNN approach for dynamics learning and the other which uses the LNN approach. We refer to them as MBRL-DNN and MBRL-LNN respectively. In both versions, we use an imagination horizon of $16$ time steps. In addition, we train a state-of-the-art model-free RL algorithm, Soft Actor-Critic (SAC)~\citep{sacv2}, to serve as a baseline. We train each algorithm on five random seeds. 

\section{Results and Analysis}
\begin{figure}[h]
    \centering
    % \captionsetup{justification=centering}
    \includegraphics[width=0.7\textwidth]{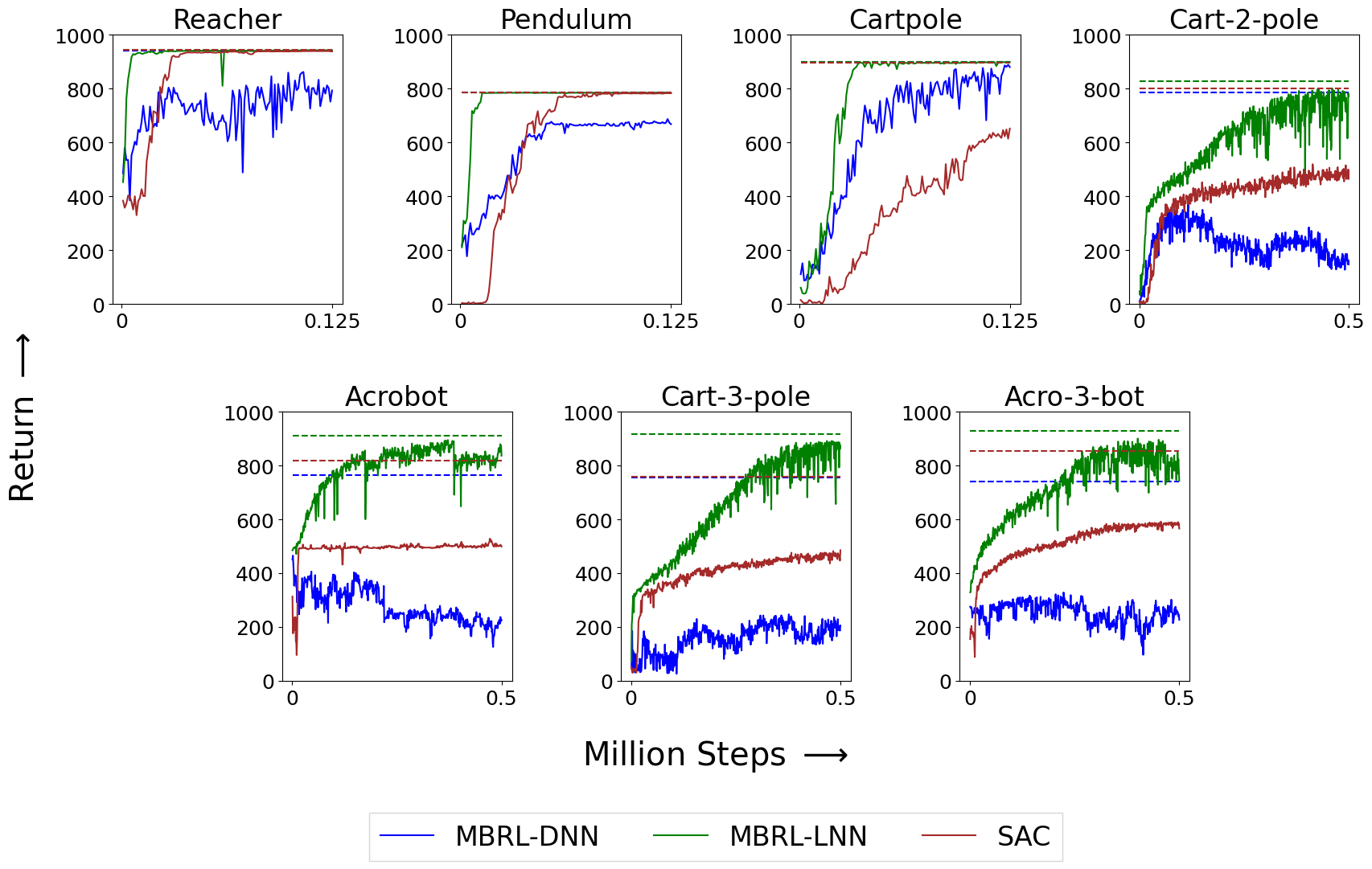}
    \caption{Training curves for MBRL-DNN, MBRL-LNN and SAC experiments. The solid curves represent the mean performance over all the seeds. The dashed lines indicate the best average-return after convergence.}
   \label{fig:results_bhvr}
\end{figure}

We record the results from the experiments in Table \ref{tab:results_table}. The training curves are shown in Figure \ref{fig:results_bhvr}. 
\begin{itemize}[leftmargin=*]
\item Across environments, MBRL-LNN achieves significantly lower dynamics error than MBRL-DNN. The reward error for both methods are similar.

\item In Reacher, Pendulum and Cartpole, all the methods, i. e., MBRL-DNN, MBRL-LNN and SAC, successfully solve the task and achieve similar average-return. In Cart-2-pole, again, all the methods are successful. However, in terms of average-return, MBRL-LNN $>$ SAC $>$ MBRL-DNN.

\item In Acrobot, Cart-3-pole and Acro-3-bot, only MBRL-LNN and SAC are successful, while MBRL-DNN is unsuccessful. Again, in terms of average-return, MBRL-LNN $>$ SAC $>$ MBRL-DNN.

\item Across environments, MBRL-LNN achieves similar or better average return than MBRL-DNN and SAC, while requiring fewer samples, i. e., MBRL-LNN achieves better sample efficiency.
\end{itemize} 

\setlength{\tabcolsep}{.3em}
   \begin{table}[h]
  \begingroup \fontsize{7.0pt}{12pt}\selectfont
  \begin{center}
  \begin{tabular}{| *{9}{c|} }
   \hline

  \multirow{2}{*}{Environment} & \multirow{2}{*}{ Method} & \multirow{2}{*}{ Steps} &  Dynamics &  Reward &  Average &  Solved &  Finite   &  Trajectory\\
  & & &  Error &  Error &   Return &   (Y/N) & Time  &  Error  \\ 
  & & & & &  &  & MLE & \\
   \hline\hline

    \multirow{3}{*}{ Reacher}  &  MBRL-DNN &  0.25 M  & 126.57e-5 & 7.2217e-4  & 942.7 & \textbf{Y} & \multirow{3}{*}{-0.0051}    & 1.0984  \\\cline{2-7}\cline{9-9}
            &  MBRL-LNN &  \textbf{0.125 M}  & \textbf{4.7557e-5} & \textbf{8.4078e-4} & \textbf{943.4} & \textbf{Y} & &  \textbf{0.2036}  \\\cline{2-7}\cline{9-9}
            &  SAC    &  1.25 M & -- & --  & 943.2 & \textbf{Y} & & -- \\\cline{2-7}\cline{9-9}
      \hline \hline

  \multirow{3}{*}{ Pendulum}  &  MBRL-DNN & 0.25 M    & 108.93e-5 &\textbf{8.198e-4} & \textbf{787.3} & \textbf{Y} & \multirow{3}{*}{-0.0811}   & 0.6933  \\\cline{2-7}\cline{9-9}
         &  MBRL-LNN & \textbf{0.125 M}  &\textbf{3.4559e-5} & 8.3962e-4  & 786.7 & \textbf{Y} & &  \textbf{0.1099}  \\\cline{2-7}\cline{9-9}
         &  SAC    & 1.25 M   & -- & --  & 786.8 & \textbf{Y} & & -- \\\cline{2-7}\cline{9-9} 
    \hline \hline

    \multirow{3}{*}{ Cartpole}  &  MBRL-DNN & 0.25 M & 149.31e-5 & \textbf{7.5226e-4}  & 899.4 & \textbf{Y}  & \multirow{3}{*}{-0.1617}   & 0.5141  \\\cline{2-7}\cline{9-9}
         &  MBRL-LNN & \textbf{0.125 M} & \textbf{6.014e-5} & 8.6215e-4  & \textbf{900.0} & \textbf{Y} & &  \textbf{0.1645}  \\\cline{2-7}\cline{9-9}
       &  SAC    & 1.25 M  & -- & --  & 897.4 & \textbf{Y} & & --   \\\cline{2-7}\cline{9-9}
      \hline \hline

    \multirow{3}{*}{ Cart-2-pole} &  MBRL-DNN & 2 M  & 499.94e-5 & 16.049e-4 & 786.4 & \textbf{Y} & \multirow{3}{*}{0.3241}  &  2.6164 \\\cline{2-7}\cline{9-9}
        &  MBRL-LNN & \textbf{0.5 M} & \textbf{70.389e-5} & \textbf{8.8135e-4} & \textbf{828.3} & \textbf{Y} & &  \textbf{0.9744} \\\cline{2-7}\cline{9-9}
        &  SAC      & 10 M  & -- & --  & 801.7 & \textbf{Y} & & --  \\\cline{2-7}\cline{9-9} 
    \hline\hline 

    \multirow{3}{*}{ Acrobot}   &  MBRL-DNN & 2 M   & 799.90e-5 & 8.848e-4 & 764.5 & N & \multirow{3}{*}{0.4265}    &  4.8975 \\\cline{2-7}\cline{9-9}
          &  MBRL-LNN & \textbf{0.5 M} & \textbf{48.046e-5} & \textbf{7.9032e-4} & \textbf{911.8} & \textbf{Y} & &  \textbf{2.2121}  \\\cline{2-7}\cline{9-9}
      &  SAC      &  10 M & --  & -- & 817.8 & \textbf{Y} & & --    \\\cline{2-7}\cline{9-9} 
   \hline\hline

    \multirow{3}{*}{ Cart-3-pole}    &  MBRL-DNN & 2 M  & 1196.0e-5 & 17.044e-4 & 757.0 & N & \multirow{3}{*}{0.8099}   &  18.2503 \\\cline{2-7}\cline{9-9}
        &  MBRL-LNN & \textbf{0.5 M}  & \textbf{61.3e-5} & \textbf{6.5141e-4} & \textbf{916.6} & \textbf{Y} & &  \textbf{1.3601}  \\\cline{2-7}\cline{9-9}
     &  SAC      & 10 M  & -- & -- & 759.4 & \textbf{Y} & & --  \\\cline{2-7}\cline{9-9} 
     \hline\hline

    \multirow{3}{*}{ Acro-3-bot}    &  MBRL-DNN & 2 M  & 1322.0e-5 & 16.6256e-4 & 739.7 & N & \multirow{3}{*}{0.6230}    &  16.0338 \\\cline{2-7}\cline{9-9}
        &  MBRL-LNN & \textbf{0.5 M}  & \textbf{156.34e-5} & \textbf{10.308e-4} & \textbf{929.4} & \textbf{Y} & &  \textbf{2.8153}  \\\cline{2-7}\cline{9-9}
     &  SAC     & 10 M  & -- & -- & 853.5 & \textbf{Y} & & --   \\\cline{2-7}\cline{9-9}
    \hline

  \end{tabular}
  \end{center}
  \endgroup
  \caption{Overall results from MBRL-DNN, MBRL-LNN and SAC experiments.}
  \label{tab:results_table}
  \end{table}

We try to understand the results better. Reacher, Pendulum and Cartpole appear to be simple environments where all methods perform well. Cart-2-pole, Acrobot, Cart-3-pole and Acro-3-bot appear to be more challenging environments, where there is a noticeable difference in the performance of different methods. We try to understand what makes these latter environments more challenging and why certain methods perform better in these environments than others.

\subsection{Imaginary Trajectories}
As a first step, we assess the quality of the imaginary trajectories produced by MBRL-DNN and MBRL-LNN in each environment. We generate 10,000 imaginary trajectories of length 16 time steps and also generate the corresponding ground truth trajectories. 

First, we assess how well the imaginary trajectories respect the underlying physical laws, such as conservation of energy. As external non-conservative forces are present (actuators), the total energy of the system will change with time. The change in total energy must equal the work done by the actuators, for energy to be conserved. We define the energy error as the absolute difference between the two quantities. For each imaginary trajectory, we compute the energy error at each time step. We then average over all the imaginary trajectories and plot the energy error as a function of time in Figure \ref{fig:imagine_energy}. We find that, across environments, MBRL-LNN achieves much lower energy error than MBRL-DNN, i. e, MBRL-LNN conserves energy much better.

\begin{center}
\begin{minipage}{\linewidth}\centering
  \begin{minipage}{0.45\linewidth}\centering
      \begin{figure}[H]
          \includegraphics[width=\linewidth]{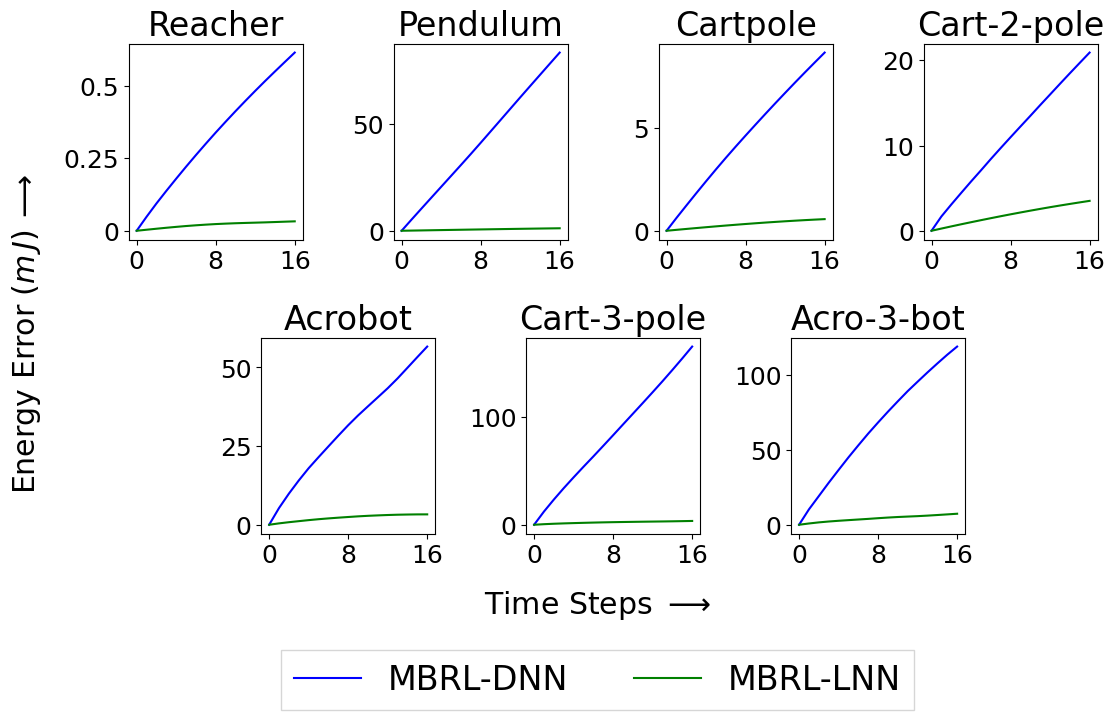}
          \captionsetup{justification=centering}
          \caption{Energy Error vs Time.}
          \label{fig:imagine_energy}
      \end{figure}
  \end{minipage}
  \hspace{0.025\linewidth}
  \begin{minipage}{0.45\linewidth}\centering
      \begin{figure}[H]
          \includegraphics[width=\linewidth]{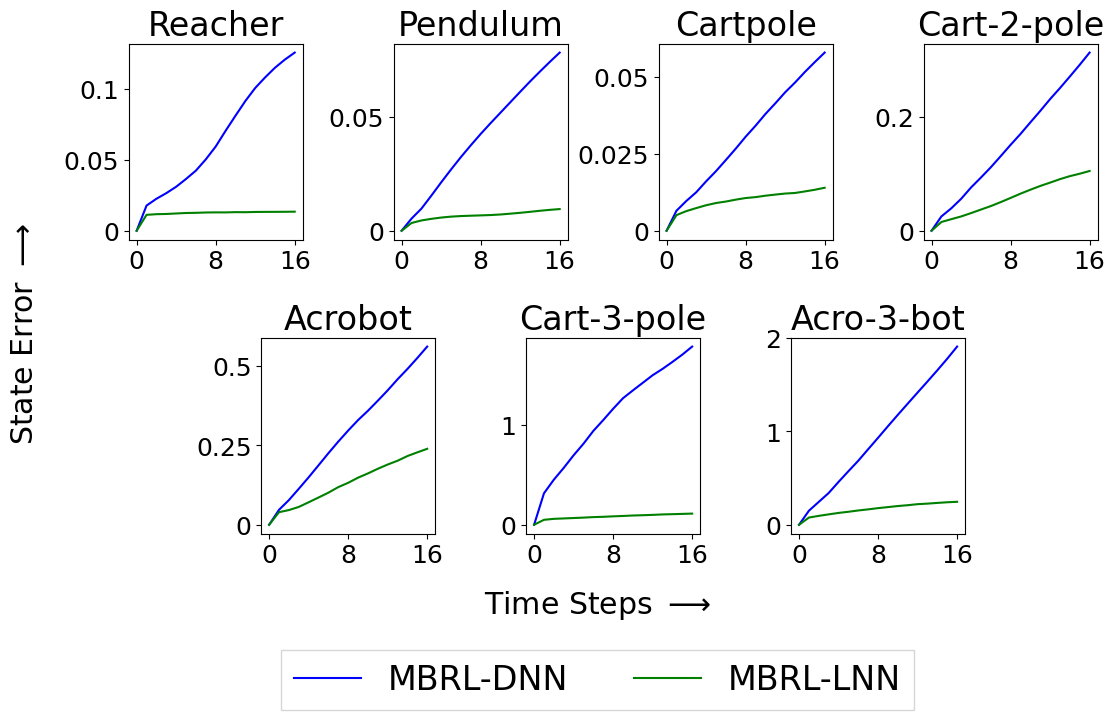}
          \captionsetup{justification=centering}          
          \caption{State Error vs Time.}
           \label{fig:imagine_state}
      \end{figure}
  \end{minipage}
\end{minipage}
\end{center}

Next, we assess the actual accuracy of the imaginary trajectories. For each imaginary trajectory, we compute the state error at each time step, i. e., the L1 error between the predicted state and the true state. We also compute the trajectory error, which we define as the sum of the state errors along the trajectory. We average both quantities over all the imaginary trajectories. We plot the state error as a function of time in Figure \ref{fig:imagine_state}. We record the trajectory error in Table \ref{tab:results_table}. We find that, MBRL-LNN has a low trajectory error across environments. Hence it is successful across environments. Whereas, MBRL-DNN has a moderate trajectory error in Reacher, Pendulum, Cartpole and Cart-2-pole, and a high trajectory error in Acrobot, Cart-3-pole and Acro-3-bot. Hence it is successful in the former environments, while it is unsuccessful in the latter environments.

\subsection{Environment Dynamics}
Next, we try to characterize the underlying dynamics of each environment by calculating their Lyapunov exponents, which measure the rate of separation of trajectories that start from nearby initial states. If the initial separation vector is $\textbf{u}_{0}$ and the separation vector at time $t$ is $\textbf{u}_{t}$, then the Lyapunov exponent is defined as $\lambda = \frac{1}{t} \log \frac{\| \textbf{u}_{t}\|}{\| \textbf{u}_{0}\|}$. The rate of separation varies based on the initial separation vector. In general, there is a spectrum of Lyapunov exponents, equal in number to the state space dimension, each associated with a direction (which changes with time). The rate of separation is maximum along the direction associated with the maximal Lyapunov exponent (MLE). An arbitrary initial separation vector will typically contain some component in this direction, and because of the exponential growth rate, this component will dominate the evolution. The MLE is a measure of the system's sensitivity to initial conditions. It also represents how quickly numerical errors will accumulate. Thus, it is also a measure of the predictability.

Typically, the MLE is computed for unforced autonomous systems of the form $\dot{\textbf{s}} = \textbf{f}(\textbf{s})$, in the long-term, i. e., as $t \rightarrow \infty$. We refer to this as the standard MLE. The simplest method to compute it is to evolve two trajectories starting from nearby initial states for a sufficiently long time and use the Lyapunov exponent definition. A more efficient and reliable method is to use the variational equation~\citep{skokos2010lyapunov}, $\dot{\textbf{u}} = \frac{d \, \textbf{f}}{d \, \textbf{s}} \, \textbf{u}$, which linearizes the dynamics to estimate how the separation vector $\textbf{u}$ evolves with time. We first consider a random initial separation vector of unit length. We then jointly integrate the system dynamics along with the variational equation to compute how the separation vector evolves with time. We renormalize the separation vector periodically and each time use the Lyapunov exponent definition to compute a fresh estimate of the MLE and update its running average. We stop this process once the running average converges. 

However, we are more interested in computing the MLE for forced autonomous systems of the form $\dot{\textbf{s}} = \textbf{f}(\textbf{s},\textbf{a})$, where $\textbf{a} \sim \boldsymbol{\pi}(. | \textbf{s})$, over the finite time period from the start of an episode, till the agent reaches the goal state, which typically takes several hundred time steps. We refer to this as the finite-time MLE. To compute it, we again follow the variational equation approach. First, we extend the variational equation to forced autonomous systems,
\begin{equation}
\dot{\textbf{u}} = \frac{d \, \textbf{f}}{d \, \textbf{s}} \, \textbf{u} = ( \frac{\partial \, \textbf{f}}{\partial \, \textbf{s}} + \frac{\partial \, \textbf{f}}{\partial \, \textbf{a}} \, \frac{d \, \textbf{a}}{d \, \textbf{s}} )\,\textbf{u} 
\end{equation}
Then, we follow the exact same procedure as earlier, except that we stop the MLE computation process once the agent reaches the goal state. The finite-time MLE depends on factors such as the system dynamics, degree of actuation, control policy and damping. In each environment, we compute the finite-time MLE for policies of MBRL-DNN, MBRL-LNN and SAC, and average the results. We record the results in Table \ref{tab:results_table}. We find that Reacher, Pendulum and Cartpole have a small, negative, finite-time MLE. This implies that they are not sensitive to initial conditions. Whereas, Cart-2-pole, Acrobot, Cart-3-pole and Acro-3-bot have a large, positive, finite-time MLE. This implies that they are sensitive to initial conditions.

\subsection{Discussion}
Cart-2-pole, Acrobot, Cart-3-pole and Acro-3-bot are underactuated (see Appendix \ref{sec:env}) and sensitive to initial conditions. Underactuation by itself makes it hard to learn successful policies. Sensitivity to initial conditions leads to high variance in the agent's actual as well as imaginary trajectories. It also leads to poor predictability, which affects critic learning, as it is concerned with predicting the expected discounted return, and hence also affects actor learning, as our methods follow an actor-critic approach. Sensitivity to initial conditions also makes it hard for DNNs to learn accurate dynamics models. Hence, these environments are challenging.  

In Acrobot, Cart-3-pole and Acro-3-bot, MBRL-DNN has a large dynamics error. As numerical errors accumulate fast in these environments, its imaginary trajectories have high trajectory error. Hence, it is unable to learn good policies and is unsuccessful in these environments. Cart-2-pole is a borderline case. Here, MBRL-DNN achieves a moderate dynamics error. Hence, even though numerical errors accumulate fast, its imaginary trajectories only have moderate trajectory error. Hence, it is able to learn a reasonably good policy and is successful.

In all four of these environments, MBRL-LNN achieves a small dynamics error due to its physics-based inductive biases. Hence, even though numerical errors accumulate fast, its imaginary trajectories have low trajectory error. Hence, it is able to learn good policies and is successful. 

SAC is successful in all four of these environments. However, MBRL-LNN achieves higher average-return than SAC in these environments. It must be noted that both methods effectively perform the same number of policy updates per episode. This implies that MBRL-LNN estimates the policy gradient more accurately than SAC. The reason behind this is that MBRL-LNN computes the policy gradient analytically, while SAC estimates it through sampling. 

Thus, MBRL-LNN achieves better average-return and sample efficiency than both MBRL-DNN and SAC in these environments.

In contrast, Reacher, Pendulum and Cartpole are not sensitive to initial conditions. In these environments, MBRL-DNN has a moderate dynamics error, MBRL-LNN has a low dynamics error and numerical errors accumulate slowly. Hence, MBRL-DNN has a moderate trajectory error, MBRL-LNN has a low trajectory error and both methods are able to learn good policies and are successful, and so is the case with SAC. In these environments, MBRL-LNN achieves similar average-return to MBRL-DNN and SAC, but achieves better sample efficiency.

\section{Conclusion}
We apply model-based RL to robotic systems undergoing rigid body motion without contacts. In our algorithm, we learn a model of the environment, essentially its transition dynamics and reward function, use it to generate imaginary trajectories and backpropagate through them to update the policy, exploiting the differentiability of the model. We compare two versions of our algorithm, one which uses a standard deep neural network based dynamics model and the other which uses a much more accurate, physics-informed neural network based dynamics model.

We show that, in model-based RL, model accuracy mainly matters in environments that are sensitive to initial conditions, where numerical errors accumulate fast. In these environments, the physics-informed version of our algorithm achieves significantly better average-return and sample efficiency. In environments that are not sensitive to initial conditions, both versions of our algorithm achieve similar average-return, while the physics-informed version achieves better sample efficiency.

The sensitivity to initial conditions depends on factors such as the system dynamics, degree of actuation, control policy and damping. We measure it using the finite-time maximal Lyapunov exponent. We compute the same using the variational equation, which linearizes the dynamics to estimate how separation vectors evolve with time. The standard variational equation is defined only for unforced autonomous systems. We extend it to forced autonomous systems.

We also show that, in challenging environments, physics-informed model-based RL achieves better average-return than state-of-the-art model-free RL algorithms such as Soft Actor-Critic, as it computes the policy-gradient analytically, while the latter estimates it through sampling.

In future work, we plan to consider both friction as well as contacts, as they are present in most real world robots. We plan to focus on robotics tasks such as manipulation and locomotion. 

\appendix
\addcontentsline{toc}{section}{Appendix}
\section*{Appendix}

\section{Environment Details} \label{sec:env}
The task to be accomplished in each environment is as follows. (1) Reacher: Control a fully actuated two-link manipulator in the horizontal plane to reach a fixed target location. (2) Pendulum: Swing up and balance a simple pendulum. (3) Cartpole: Swing up and balance an unactuated pole by applying forces to a cart at its base. (4) Cart-2-pole: One extra pole is added to Cartpole. Only the cart is actuated. (5) Acrobot: Control a two-link manipulator to swing up and balance. Only the second pole is actuated. (6) Cart-3-pole: One extra pole is added to Cart-2-pole. Only the cart and the third pole are actuated. (7) Acro-3-bot: One extra pole is added to Acrobot. Only the first and the third poles are actuated. In all the environments, there are no terminal states. Each episode consists of $1000$ time steps. The total reward over an episode is in the range $[0,1000]$.

\section{Implementation Details }
Table \ref{tab:hyperparams_mbrl} and Table \ref{tab:arch_mbrl} list the hyperparameters and network architectures used in our model-based RL experiments. For SAC, we use the same hyperparameters and network architectures as the original SAC paper~\citep{sacv2}.
\begin{table}[h]
\centering
\begingroup \fontsize{8pt}{10pt}\selectfont

\begin{minipage}{0.55\textwidth}\centering
    \begin{tabular}{||c c||} 
     \hline
     Parameter & Value\\
     \hline
     Random episodes at start of training ($K$) & 10\\
     Replay buffer size & $10^{5}$\\
     Batch size for model learning & $64$\\
     Model learning batches per episode ($N_{1}$) & $10^{4}$\\
     Batch size for behaviour learning & $64$\\
     Behaviour learning batches per episode ($N_{2}$) & $10^{3}$\\
     Imagination horizon ($T$) & 16\\
     Discount factor ($\gamma$) & $0.99$\\
     Lambda ($\lambda$) & $0.95$\\
     Entropy weightage ($\eta$) & $10^{-4}$\\
     Gradient clipping norm & $100$\\
     Optimizer & AdamW\\
     Learning rate & $3\times10^{-4}$\\
     \hline
    \end{tabular}
    \captionsetup{justification=centering}
    \caption{Model-Based RL Hyperparameters}
    \label{tab:hyperparams_mbrl}
\end{minipage}
\begin{minipage}{0.425\textwidth}\centering
    \begin{tabular}{||c c||} 
     \hline
     Network & Architecture\\
     \hline
    Critic & $256$ , $256$ , $1$\\
    Actor & $256$ , $256$ , $2 \times \dim\,(a)$\\
    DNN & $64$ , $64$ , $\dim\,(s)$\\
    LNN L & $64$ , $64$ , $\frac{\dim\,(s) \times (\dim\,(s)+2)}{8}$\\
    LNN V & $64$ , $64$ , $1$\\
    Reward & $64$ , $64$ , $1$\\
     \hline
    \end{tabular}
    \captionsetup{justification=centering}
    \caption{Model-Based RL \\Network Architectures}
    \label{tab:arch_mbrl}
\end{minipage}

\endgroup
\end{table}

\section{Project Webpage and Code}

Project Webpage : \href{https://adi3e08.github.io/research/pimbrl}{https://adi3e08.github.io/research/pimbrl}

\noindent Code : \href{https://github.com/adi3e08/Physics_Informed_Model_Based_RL}{https://github.com/adi3e08/Physics\_Informed\_Model\_Based\_RL}

\clearpage
\bibliography{main}

\end{document}